\documentclass[letterpaper]{article} 
\usepackage{aaai2026}  
\usepackage{times}  
\usepackage{helvet}  
\usepackage{courier}  
\usepackage[hyphens]{url}  
\usepackage{graphicx} 
\usepackage{natbib}  
\usepackage{caption} 
\usepackage{longtable}
\usepackage{algorithmic}
\usepackage{microtype}
\usepackage{subfigure}
\usepackage{booktabs}
\usepackage{amssymb}
\usepackage{mathtools}
\usepackage{amsthm}
\usepackage{soul}
\usepackage{longtable}
\usepackage{multirow}
\usepackage{xspace}
\usepackage{pifont}
\usepackage{latexsym}
\usepackage{inconsolata}
\usepackage{arydshln}
\usepackage{enumitem}
\usepackage{array}
\usepackage{utfsym}
\usepackage{listings}
\usepackage[dvipsnames]{xcolor}

\theoremstyle{plain}
\newtheorem{theorem}{Theorem}

\theoremstyle{definition}

\theoremstyle{remark}

\usepackage[table]{xcolor}  
\usepackage{pifont}

\definecolor{shadecolor}{rgb}{.92, .92, .92}
  {\endMakeFramed}

\usepackage[ruled,vlined,linesnumbered]{algorithm2e}
\usepackage{tcolorbox}
\tcbuselibrary{listingsutf8}

\definecolor{teaserred}{RGB}{180,10,56}

\definecolor{teaserblue}{RGB}{0,15,139}

\definecolor{uclablue}{RGB}{159, 195, 224}

\definecolor{uclagold}{RGB}{254,180,167}

\definecolor{grayred}{RGB}{232,237,205}

\definecolor{TealBlue}{rgb}{1.0, 0.97, 0.8}

\definecolor{iccvblue}{rgb}{0.16, 0.54, 0.86}
\newtcolorbox{ttcolorbox}[1][]{
  colframe=iccvblue,
  colback=iccvblue!5!white,
  title=#1,
  fonttitle=\bfseries\sffamily,
}

\definecolor{MyCyan}{RGB}{0,163,218}
\definecolor{MyDarkBlue}{RGB}{0,103,165}
\definecolor{MyDarkGreen}{RGB}{56,116,51}
\definecolor{MyMagenta}{RGB}{200,18,126}
\definecolor{iccvblue}{rgb}{0.21,0.49,0.74}
\definecolor{lightyellow}{RGB}{255,249,207}

\title{
InEx: Hallucination Mitigation via \\ Introspection and Cross-Modal Multi-Agent Collaboration}

\author{
    Zhongyu Yang\textsuperscript{\rm 1,2$^*$}, Yingfang Yuan\textsuperscript{\rm 2$^*$},
    Xuanming Jiang\textsuperscript{\rm 1$\dagger$},  
    Baoyi An\textsuperscript{\rm 1}, 
    Wei Pang\textsuperscript{\rm 2$\dagger$}
    \\
}

\affiliations{
    \textsuperscript{\rm 1}Xi’an Jiyun Technology Co., Ltd., Xi’an, China  \textsuperscript{\rm 2}BCML, Heriot-Watt University, Edinburgh, UK\\
    \{zy4028, y.yuan, w.pang\}@hw.ac.uk, jiangxm24@stu.xjtu.edu.cn, anby20@lzu.edu.cn
}

\definecolor{mydarkblue}{rgb}{0,0.08,0.45}
\definecolor{darkgreen}{rgb}{0.0, 0.5, 0.0} 
\definecolor{myblue}{RGB}{235,235,250}
\definecolor{lightpink}{RGB}{225, 235, 250}
\definecolor{lightblue}{RGB}{230, 235, 245}  
\definecolor{lightgray}{RGB}{240, 240, 240}  
\definecolor{darkgray}{RGB}{220, 220, 220} 

\definecolor{superlightred}{rgb}{0.99, 0.92, 0.92}
\definecolor{darkgreen}{RGB}{50,100,0}
\definecolor{darkred}{RGB}{230, 150, 170}

\newcommand{\up}[1]{{\footnotesize \ $\uparrow${#1}}}
\newcommand{\downbad}[1]{{\footnotesize \ $\downarrow${#1}}}
\newcommand{\down}[1]{{\footnotesize \ $\downarrow${#1}}}

\newcommand{\basex}[1]{\textcolor{gray!85}{\footnotesize \ $\uparrow${#1}}}

\usepackage{bibentry}

\begin{document}
\maketitle
\footnotetext[1]{$^*$ Equal contribution}
\footnotetext[2]{$\dagger$ Corresponding Author}

\begin{abstract}
Hallucination remains a critical challenge in large language models (LLMs), hindering the development of reliable multimodal LLMs (MLLMs).
Existing solutions often rely on human intervention or underutilize the agent’s ability to autonomously mitigate hallucination.
To address these limitations, we draw inspiration from how humans make reliable decisions in the real world. They begin with introspective reasoning to reduce uncertainty and form an initial judgment, then rely on external verification from diverse perspectives to reach a final decision.
Motivated by this cognitive paradigm, we propose InEx, a training-free, multi-agent framework designed to autonomously mitigate hallucination.
InEx introduces internal introspective reasoning, guided by entropy-based uncertainty estimation, to improve the reliability of the decision agent’s reasoning process. The agent first generates a response, which is then iteratively verified and refined through external cross-modal multi-agent collaboration with the editing agent and self-reflection agents, further enhancing reliability and mitigating hallucination.
Extensive experiments show that InEx consistently outperforms existing methods, achieving 4\%–27\% gains on general and hallucination benchmarks, and demonstrating strong robustness. 
\end{abstract}

\section{Introduction}
\label{intro}
Large language models (LLMs) have enabled substantial progress in natural language processing, leading to the development of multimodal LLMs (MLLMs) capable of understanding both visual and textual information~\cite{zhang2025openmmreasonerpushingfrontiersmultimodal, LongVA, yang2025longvtincentivizingthinkinglong}. 
Despite their remarkable capabilities, MLLMs face the critical and unresolved challenge of hallucination. 
This phenomenon refers to generating responses that, while linguistically plausible, are factually inaccurate relative to the input image.
Hallucination significantly compromises the reliability of MLLMs and presents a substantial barrier to their safe and effective deployment in real-world applications~\cite{mermaid}.

Hallucination mitigation has been approached through three primary paradigms, each constrained by specific limitations that collectively motivate our work.
Pre-processing methods, such as fine-tuning or reinforcement learning from human feedback~\cite{perturbollava,rlhf_v,rlaif_v}, require substantial human intervention, including supervision and task-specific annotations, thus limiting scalability and adaptability.
In-processing methods guide generation using hallucination-related heuristics or metrics~\cite{farquhar2024detecting,zhang2025uncertainty}. However, because these methods are tightly coupled with the model’s internal reasoning and lack external verification, they remain vulnerable to meta-hallucination (i.e., meta-cognitive confidence in claims derived from incorrect knowledge) due to the model’s inherent stereotype biases.
Post-processing methods often employ an external verifier to detect or correct errors after generation~\cite{Madaan2023SelfRefineIR, wu2025reverse}. In some cases, these verifiers require retraining, incurring additional computational overhead. Furthermore, we argue that treating in-processing and post-processing methods in isolation ultimately underutilizes the autonomous capabilities of agents for hallucination mitigation. In real-world decision-making, iterative reasoning and verification are inherently interconnected processes that mutually reinforce one another.

To address these challenges, we draw inspiration from the cognitive paradigm that humans make reliable decisions. This process typically begins with introspective internal reasoning, forming a preliminary judgment by reducing uncertainty. This decision is then verified and refined through external feedback from diverse perspectives, ultimately leading to a well-grounded decision. Motivated by this observation, we propose \textbf{InEx}, a training-free multi-agent framework designed to autonomously mitigate hallucination.

In InEx, a decision agent leverages unsupervised uncertainty estimation to perform internal introspective reasoning (\textbf{In}), proactively retrieving and analyzing visual information to generate a response without human intervention.  
To further mitigate hallucination, InEx introduces external cross-modal multi-agent collaboration (\textbf{Ex}) for iterative verification and refinement. The initial response from the decision agent undergoes repeated evaluation by self-reflection agents and refinement by the decision agent, guided by cross-modal evidence.
This process of alternating verification and introspective refinement continues until the response achieves cross-modal consensus or meets a predefined stopping criterion, ensuring autonomy and reliability in response generation.


InEx unifies In and Ex into a cohesive process. Their interaction enables proactive uncertainty reduction and iterative cross-modal verification and refinement, with each component reinforcing the effectiveness of the other.
Together, these modules enable fully autonomous hallucination mitigation, requiring neither human supervision nor model retraining.
As a result, InEx consistently outperforms existing baselines across both general-purpose and hallucination benchmarks, while delivering well-calibrated and trustworthy responses.


Our main contributions can be summarized as follows:

\begin{itemize}
    \item We propose \textbf{InEx}, a training-free multi-agent framework that mitigates hallucination via internal introspective reasoning and external cross-modal collaboration.

    \item  We demonstrate that cross-modal consensus can effectively mitigate hallucination with the support of multi-modal, multi-perspective self-reflection agents.

    \item Extensive experiments show that InEx consistently outperforms prior baselines by 3.8\%–16.2\% on general-purpose benchmarks and 6.5\%–26.7\% on hallucination benchmarks, confirming its effectiveness and generalization.
\end{itemize}
\section{Related Works}
\label{relatedwork}

\noindent \textbf{Hallucination Mitigation in MLLMs:}  
Prior work can be categorized into three main paradigms, each with inherent limitations:
\textbf{(i) Data-centric tuning} leverages contrastive learning or reinforcement learning on curated adversarial prompts or hard negatives~\citep{hacl,rit,gfaif}.
While effective on specific tasks, such methods incur high annotation costs and demand task-specific retraining, significantly hindering scalability and generalization.
\textbf{(ii) Post-hoc revision} modifies generated outputs using external verifiers, regenerators, or retrieval-augmented modules~\citep{Pelican, woodpecker,degf,amoh}.
However, these methods are reactive rather than preventive, leading to latency, fragile pipelines, and ultimately failing to address the root causes of hallucination.
\textbf{(iii) Decoding-time control} regulates token-level generation through logit calibration or uncertainty-based penalties~\citep{halc,gcd,sid,yang2025script}.
Despite their efficiency, these techniques often generate overconfident yet unreliable outputs due to shallow, modality-unaware reasoning and poor uncertainty modeling.
Our \textbf{InEx} bridges this gap by unifying proactive, modality-aware uncertainty modeling with structured cross-modal verification.

\noindent\textbf{Self-reflection:}  
Recent studies have explored self-reflection mechanisms in LLMs, in which models iteratively refine their outputs through internal feedback loops to enhance reasoning and interpretability.
These approaches can be categorized into three types:
(1) self-evaluation within the textual modality~\cite{Shinn2023ReflexionLA, Renze_2024};
(2) feedback from an external trained textual critic~\cite{Gou2023CRITICLL, wikiautogen}; and
(3) integration of external knowledge sources such as Wikipedia or web search~\cite{Madaan2023SelfRefineIR, Asai2023SelfRAGLT}. 
Although effective for text-only reasoning, these approaches lack visual grounding, limiting their applicability to multimodal contexts.
To overcome this limitation, \textbf{InEx} extends self-reflection to the multimodal domain through structured collaboration among complementary visual and textual agents, enabling peer-review-style verification that enhances reasoning reliability.

\begin{figure}[!ht]
\centering
\includegraphics[width=\linewidth]{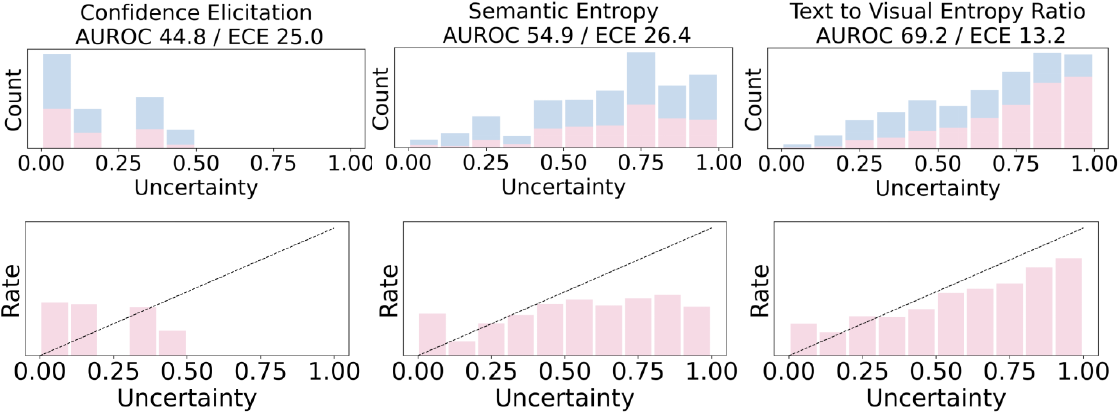}
\caption{
{\colorbox[HTML]{F5D6E3}{\phantom{xx}} Hallucination \quad
 \colorbox[HTML]{C2D6EC}{\phantom{xx}} No~Hallucination} \\
Uncertainty calibration across different methods, with each column showing uncertainty score distributions (top) and corresponding hallucination rates (bottom) across bins.
}
\label{fig:calibration}
\end{figure}

\begin{figure*}[ht]
\centering
\includegraphics[width=\linewidth]{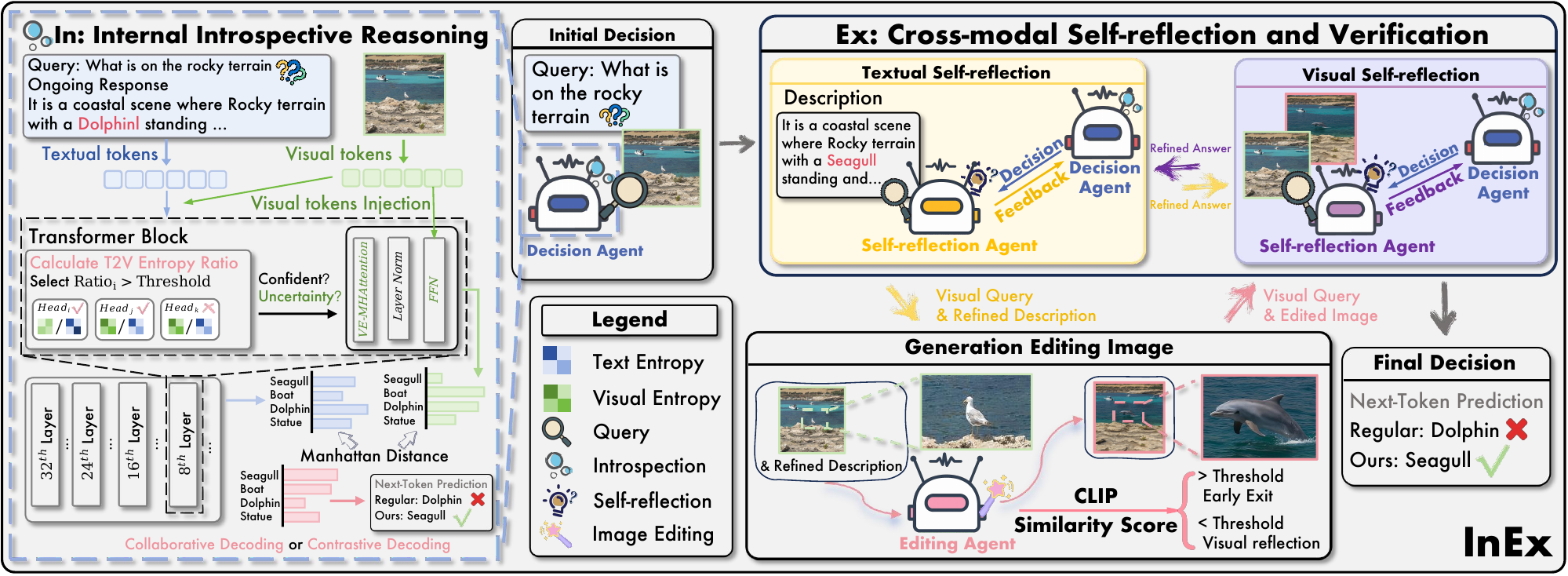}
\caption{Overview of \textbf{InEx}. \textbf{In:} A decision agent initiates introspective reasoning guided by unsupervised uncertainty estimation, producing an initial response grounded in internal uncertainty signals. \textbf{Ex:} The response is then iteratively refined through alternating cross-modal verification and introspective updates, where self-reflection agents assess consistency with visual and textual evidence, and the decision agent continues to revise its output accordingly.}
\label{fig:pipeline}
\end{figure*}

\section{Preliminaries}

\noindent \textbf{MLLM Architecture:}
An MLLM $f_\theta$, parameterized by $\theta$, generates a response sequence $\mathbf{y} =[y_1, \dots, y_T]$, where each $y_t$ is a text token, conditioned on an input image $v$ and textual query $\mathbf{x}$.
The image $v$ is first processed by a vision encoder, then transformed into a set of visual tokens $\mathbf{z} = [\mathbf{z}_1, \dots, \mathbf{z}_M]$ via a vision-language alignment module. These visual tokens, along with the query tokens from $\mathbf{x}$, are jointly fed into an LLM to generate the response $\mathbf{y}$. 
The LLM adopts a standard decoder-style architecture with \( L \) transformer layers. To predict token \( y_t \in \mathbf{y} \), the hidden state is updated sequentially layer by layer. Given the hidden representation \( \mathbf{H}_t^{\ell-1} \) from layer \( \ell-1 \), the forward update at layer \( \ell \) is defined as follows:
\begin{small}
\begin{equation}
\label{eq.mha}
\bar{\mathbf{H}}_{t}^\ell = \operatorname{MHA}_\ell(\mathbf{H}_t^{\ell-1}) + \mathbf{H}_t^{\ell-1}, \,
\mathbf{H}_t^{\ell} = \operatorname{FNN}_\ell(\bar{\mathbf{H}}_t^\ell) + \bar{\mathbf{H}}_t^\ell, 
\end{equation}
\end{small}
\noindent where $\operatorname{MHA}_\ell(\cdot)$ and $\operatorname{FNN}_\ell(\cdot)$ denote the multi-head self-attention and feed-forward network at layer \( \ell \), respectively.
Each $\operatorname{MHA}_\ell$ module consists of \( H \) attention heads, and is computed as:
$\operatorname{MHA}_\ell\left(\mathbf{H}_t^{\ell-1}\right) = \operatorname{Concat}(\text{Head}_{\ell, 1}, \dots, \text{Head}_{\ell, H}) \mathbf{W}_\ell^o$,
where $\mathbf{W}_\ell^o$ is a learnable parameter matrix at layer \( \ell \).

A projection head $\phi (\cdot)$ predicts the logits of the token $y_t$: $\phi\left(\mathbf{H}_t^L\right) = \mathbf{H}_t^L\mathbf{W}_{\text {out}} + \mathbf{b}.$
Finally, the model produces the probability distribution over the vocabulary as:
$ p_\theta\left(y_t \mid v, \mathbf{x}, \mathbf{y}_{<t}\right)=\operatorname{softmax}\left(\phi\left(\mathbf{H}_t^{L}\right)\right)_{y_t}.$

\noindent \textbf{Text-to-Visual Entropy Ratio (TVER):} 
A number of methods~\cite{farquhar2024detecting, xiong2024can, zhang2025uncertainty, wan2025only} detect hallucination by uncovering patterns in the model’s internal reasoning process. In this work, we systematically investigate these methods and empirically find that TVER is the most effective, achieving the highest AUROC and lowest ECE (Figure~\ref{fig:calibration}), thereby indicating superior calibration and a strong correlation with hallucination. 

For each attention head at layer $\ell$, TVER first computes the raw attention distribution as follows:
\begin{equation}
\mathbf{a}_{\ell} = \operatorname{softmax}\left(Q_{\ell} K_{\ell}^\top / \sqrt{d_k}\right).
\end{equation}
The $\mathbf{a}_{\ell}$ are partitioned based on textual indices $\mathcal{T}$ and visual indices $\mathcal{V}$, and the entropy for each subset is computed as shown in Eq.~\ref{eq:entropy}, where $i \in \mathcal{T}$ or $i \in \mathcal{V}$, and $\boldsymbol{p}_{\ell,i} = \operatorname{softmax}(\mathbf{a}_{\ell,i})$.
\begin{equation}
\label{eq:entropy}
\operatorname{Entropy}(\mathcal{T} || \mathcal{V}) = -\sum_{|\mathcal{T} || \mathcal{V}|} \boldsymbol{p}_{\ell,i} \log \boldsymbol{p}_{\ell,i},   
\end{equation}
The TVER at layer $\ell$ for an attention head is defined as:
\begin{small}
\begin{equation}
\label{eq:tverratio}
\mathrm{TVER}_{\ell} = \frac{\operatorname{Entropy}(\mathcal{T})}{\operatorname{Entropy}(\mathcal{V})}.
\end{equation}  
\end{small}

As shown in Figure~\ref{fig:calibration}, a higher TVER score indicates a greater likelihood of hallucination, corresponding to increased entropy in $\mathcal{T}$ and decreased entropy in $\mathcal{V}$. Since entropy reflects uncertainty, lower entropy in $\mathcal{V}$ suggests confidence in the visual information. In such cases, elevated uncertainty in the textual modality may stem from misleading yet confident visual cues. To guide reasoning and mitigate hallucination, attention heads $h \in H$ with TVER scores above the threshold $\gamma_{\text{TVER}}$ are masked as follows:
\begin{equation}
\tilde{\mathbf{a}}_{\ell, h} = 
\begin{cases}
0, & \mathrm{TVER}_{\ell} \geq \gamma_\text{{TVER}}, \\
\mathbf{a}_{\ell, h}, & \text{otherwise}.
\end{cases}
\end{equation}
Finally, we propose the Vision-Enhanced Multi-Head Attention (VE-MHA) as follows:

\begin{equation}
\label{eq:ve-mha}
\operatorname{VE-MHA}(\bar{\mathbf{H}}_t^\ell) = \operatorname{Concat}(\tilde{\mathbf{a}}_{\ell,1}\bar{\mathbf{H}}^{\ell}_{1}, \dots, \tilde{\mathbf{a}}_{\ell,H}\bar{\mathbf{H}}^{\ell}_{ H})\mathbf{W}^o_\ell,
\end{equation}
where $\bar{\mathbf{H}}^{\ell}_{t}$ denotes the output of $\operatorname{MHA}(\cdot)$ as defined in Eq.~\ref{eq.mha}, and $[\bar{\mathbf{H}}_1^{\ell}, \dots, \bar{\mathbf{H}}_H^{\ell}]$ represents the outputs from individual original attention heads. $\operatorname{VE-MHA}$ is applied after the original multi-head attention layer to refine its output through Self-Introspective Visual Augmentation, Enhanced Logits, and Decoding, which are discussed in the following section.

\section{Methodology}
\label{method}
Our proposed \textbf{InEx} framework mitigates hallucination from two complementary perspectives: \textbf{In}, which enhances internal reasoning via TVER-guided introspection, and \textbf{Ex}, which conducts external cross-modal multi-agent collaboration.

\subsection{In: Internal Introspective Reasoning}
\label{sec:in_stage}
As shown in Figure~\ref{fig:pipeline} and Algorithm~\ref{alg:inex}, the MLLM decision agent $\mathcal{A}_d$ generates a response $\mathbf{y}$ conditioned on both visual and textual inputs.
In real-world decision-making scenarios, humans typically engage in internal reasoning. This process is supported by introspection, which is guided by empirically validated cues to avoid errors and reduce uncertainty. 
Inspired by this cognitive mechanism, we introduce TVER as an introspective signal for $\mathcal{A}_d$, triggering an internal refinement process to improve decision reliability.
When $\mathcal{A}_d$ encounters ambiguous or underdetermined contexts (i.e., when $\text{TVER}_\ell \geq \gamma{_\text{TVER}}$), it engages in self-introspective internal reasoning composed of three components: self-introspective augmentation, self-introspective enhanced logits, and self-introspective decoding. These mechanisms work together to reinforce relevant visual information, recalibrate confidence levels, and refine the response generation process. For simplicity, in the following paragraphs, we omit the subscript $t$ for hidden representations $\mathbf{H}$. By default, each $\mathbf{H}$ is assumed to correspond to a token $y_t$.

\noindent \textbf{Self-Introspective Visual Augmentation:}
As indicated by $\text{TVER}_\ell \geq \gamma{_\text{TVER}}$, $\mathcal{A}_d$ proactively retrieves relevant visual information to augment its reasoning trajectory through a memory-style modulation of the feed-forward network ($\operatorname{FFN}$) at layer $\ell$. Given a hidden state \( \bar{\mathbf{H}}^{\ell} \) as shown in Eq.~\ref{eq.mha} and visual tokens \( \mathbf{z} = \{\mathbf{z}_{1}, \dots, \mathbf{z}_{|\mathbf{z}|} \} \), we at first define a similarity-weighted retrieval: $\Delta(\mathbf{z} \mid \bar{\mathbf{H}}^{\ell} ) = \sum_{i=1}^{|\mathbf{z}|} \sigma(\langle \bar{\mathbf{H}}^{\ell} , \mathbf{z}_{i} \rangle) \cdot \mathbf{z}_{i},$
\noindent 
where \( \sigma(\cdot) \) is an activation function (e.g., ReLU or SiLU). To incorporate retrieval into the reasoning process and augment it, the $\operatorname{FFN}(\cdot)$ in Eq.~\ref{eq.mha} is redefined as an enhanced $\operatorname{FFN}(\cdot)$:
$\operatorname{FFN}_{\ell}^{\text{introspect}}(\bar{\mathbf{H}}^{\ell} ) = \alpha \Delta(\mathbf{z} \mid \bar{\mathbf{H}}^{\ell} ) + (1 - \alpha)\operatorname{FFN}_{\ell}(\bar{\mathbf{H}}^{\ell}),$
where \( \alpha \in [0, 1] \) reflects the introspection strength. Through this process, the model introspectively ``revisits'' its visual grounding to recalibrate its internal semantic space, thereby reinforcing alignment between modalities under uncertainty.

\noindent \textbf{Self-Introspective Enhanced Logits:}
Furthermore, since the logits play a key role in determining the generated response, we incorporate $\operatorname{VE\text{-}MHA}$ at the final Transformer layer $L$ to enhance the reliability of the information used in computing the logits. This is achieved by filtering out uncertain hidden representations while retaining the visual evidence introduced through $\operatorname{FFN}_{\ell}^{\text{introspect}}$. The computation of the hidden representation used for generating the logits is defined by Eqs.~\ref{eq:lasttransfomermha}-\ref{eq:lasttransfomerfnn}. The masked heads are determined at layer $\ell$. Given the introspective representation $\hat{\mathbf{H}}^L$, the corresponding enhanced logits are defined as $\phi(\hat{\mathbf{H}}^L)$.
\begin{equation}
\label{eq:lasttransfomermha}
\bar{\mathbf{H}}^{L}=\operatorname{MHA}_{L}\left(\mathbf{H}^{L-1}\right)+\mathbf{H}^{L-1},
\end{equation}
\begin{equation}
\label{eq:lasttransofmervemha}
\mathbf{H}^{L} = \operatorname{VE-MHA}_{L}(\bar{\mathbf{H}}^{L})+\bar{\mathbf{H}}^L,
\end{equation}
\begin{equation}
\label{eq:lasttransfomerfnn}
\hat{\mathbf{H}}^L = \operatorname{FNN}_{L}(\mathbf{H}^{L}) + \mathbf{H}^{L}.
\end{equation}

\noindent \textbf{Self-Introspective Decoding:}
In the real world, when new evidence contradicts prior conclusions, humans typically compare the alternatives and adopt an appropriate strategy to reach a final decision. Inspired by this behavior, we propose self-introspective decoding to reconcile the original and enhanced predictions. Specifically, we compute the Manhattan distance between the two logits for each generated text token $y_t$ using $d_t =  \|  \phi\left({\mathbf{H}}^L\right) - \phi\left(\hat{\mathbf{H}}^L\right)  \|_T
$. To improve efficiency, the distance $d_t$ is typically calculated over the top-$k$ logits (e.g., $k=20$). The final fused logits are obtained as:
\begin{small}
\begin{equation}
\text{Fused logits} = 
\begin{cases}
\phi\left(\mathbf{H}^L\right) + \alpha_1 \phi\left(\hat{\mathbf{H}}^L\right), & \text{if } d_t < \gamma_\text{d} , \\
(1+\alpha_2)\phi\left(\mathbf{H}^L\right) - \alpha_2 \phi\left(\hat{\mathbf{H}}^L\right), & \text{if } d_t \geq \gamma_\text{d},
\end{cases}
\end{equation}    
\end{small}
where \( \gamma_d \) is a confidence threshold, and \( \alpha_1, \alpha_2 \) are weight factors. When $d_t < \gamma_{{d}}$, the original and enhanced logits are similar, indicating agreement. We apply a collaborative mode by combining both representations to reinforce the decision. When $d_t \geq \gamma_{{d}}$, the predictions diverge, signaling potential conflict. We then use a contrastive mode, emphasizing the original logits and using the enhanced ones only to calibrate them. Finally, the fused logits are used with $\operatorname{softmax}$ to generate $y_t$ in an autoregressive manner.


\begin{table*}[ht]
\centering
\belowrulesep=0pt
\aboverulesep=0pt
\vspace{-2.5mm}
\resizebox{1.0\linewidth}{!}{
\renewcommand{\arraystretch}{1.2}
\begin{tabular}{l|lcccccccccc}
\toprule
\multirow{2}{*}[-0.5ex]{{\textbf{Evaluation}}} 
 & \multirow{2}{*}[-0.5ex]{{\textbf{Methods}}} 
 & \multicolumn{2}{c}{\;\textbf{Random}}  
 & \multicolumn{2}{c}{\;\textbf{Popular}} 
 & \multicolumn{2}{c}{\;\textbf{Adversarial}} 
 & \multicolumn{2}{c}{\;\textbf{Average}} \\ 
\cmidrule(l){3-4} \cmidrule(l){5-6} \cmidrule(l){7-8} \cmidrule(l){9-10}
&  
& Accuracy $\uparrow$ & F1-score $\uparrow$ 
& Accuracy $\uparrow$ & F1-score $\uparrow$  
& Accuracy $\uparrow$ & F1-score $\uparrow$  
& Accuracy $\uparrow$ & F1-score $\uparrow$  
\\ 
\midrule
\multirow{2}{*}[-4ex]{{\textbf{MSCOCO}}}
 & {\textemdash} 
 & 83.49 \basex{0.0} & 82.28 \basex{0.0} 
 & 79.98 \basex{0.0} & 79.34 \basex{0.0} 
 & 76.03 \basex{0.0} & 76.26 \basex{0.0} 
 & 79.83 \basex{0.0} & 79.29 \basex{0.0} 
\\

& OPERA \citep{opera2024cvpr}
 & \underline{87.63} \up{4.2} & 86.45 \up{4.2} 
 & \underline{83.91} \up{4.0} & \underline{83.50} \up{4.1}
 & 80.88 \up{4.9} & \underline{80.69} \up{4.4}
 & \underline{84.14} \up{4.3} & \underline{83.55} \up{4.3} 
\\

& ICD \citep{wang2024mitigating}
 & 84.97 \up{1.5} & 83.23 \up{1.0} 
 & 82.83 \up{2.9} & 81.39 \up{2.1} 
 & \underline{81.17} \up{5.1} & 80.06 \up{3.7} 
 & 82.97 \up{3.1} & 81.56 \up{2.3} 
\\

& VCD \citep{vcd2024cvpr}
 & 86.84 \up{3.4} & \underline{86.83} \up{4.6} 
 & 82.65 \up{2.7} & 83.37 \up{4.0} 
 & 78.31 \up{2.3} & 79.28 \up{3.0} 
 & 82.60 \up{2.7} & 83.16 \up{3.9} 
\\

& {\textbf{InEx (ours)}} 
 & \sethlcolor{lightpink}\hl{\textbf{89.48}} \up{6.0}  
 & \sethlcolor{lightpink}\hl{\textbf{88.24}} \up{6.0}
 & \sethlcolor{lightpink}\hl{\textbf{88.77}} \up{8.7}  
 & \sethlcolor{lightpink}\hl{\textbf{87.72}} \up{8.4}
 & \sethlcolor{lightpink}\hl{\textbf{85.97}} \up{9.9}  
 & \sethlcolor{lightpink}\hl{\textbf{85.68}} \up{9.4}
 & \sethlcolor{lightpink}\hl{\textbf{88.73}} \up{8.9}  
 & \sethlcolor{lightpink}\hl{\textbf{86.48}} \up{7.2}
\\ \cmidrule(r){1-2}
\multirow{2}{*}[-4ex]{{\textbf{A-OKVQA}\quad}}
& {\textemdash}
 & 83.45 \basex{0.0} & 82.56 \basex{0.0} 
 & 79.90 \basex{0.0} & 79.59 \basex{0.0}
 & 74.04 \basex{0.0} & 75.15 \basex{0.0} 
 & 79.13 \basex{0.0} & 79.10 \basex{0.0} 
\\

& OPERA \citep{opera2024cvpr}\;
 & \underline{88.32} \up{4.9} & \underline{87.64} \up{5.1}  
 & \underline{85.59} \up{5.7} & \underline{84.64} \up{5.1} 
 & \underline{79.17} \up{5.1} & \underline{79.97} \up{4.8} 
 & \underline{84.36} \up{5.2} & \underline{84.08} \up{5.0}
\\

& ICD \citep{wang2024mitigating}
 & 85.41 \up{2.0} & 84.99 \up{2.6} 
 & 81.79 \up{1.9} & 81.98 \up{2.4} 
 & 77.73 \up{3.7} & 78.79 \up{3.2}
 & 81.64 \up{2.5} & 82.00 \up{2.8}
\\

& VCD \citep{vcd2024cvpr}
 & 86.15 \up{2.7} & 86.34 \up{3.8} 
 & 81.85 \up{2.0} & 82.82 \up{3.2} 
 & 74.97 \up{0.9} & 77.73 \up{2.6} 
 & 80.99 \up{1.9} & 82.30 \up{3.2}
\\

& {\textbf{InEx (ours)}}
 & \sethlcolor{lightpink}\hl{\textbf{92.63}} \up{9.2}  
 & \sethlcolor{lightpink}\hl{\textbf{91.18}} \up{8.6}
 & \sethlcolor{lightpink}\hl{\textbf{88.41}} \up{8.5}  
 & \sethlcolor{lightpink}\hl{\textbf{87.99}} \up{8.4}
 & \sethlcolor{lightpink}\hl{\textbf{80.91}} \up{6.9}  
 & \sethlcolor{lightpink}\hl{\textbf{82.86}} \up{7.7}
 & \sethlcolor{lightpink}\hl{\textbf{87.32}} \up{8.2}  
 & \sethlcolor{lightpink}\hl{\textbf{87.34}} \up{8.3}
\\ \cmidrule(r){1-2}

\multirow{2}{*}[-4ex]{{\textbf{GQA}}}
& {\textemdash}
 & 83.73 \basex{0.0} & 82.95 \basex{0.0}
 & 78.17 \basex{0.0} & 78.37 \basex{0.0}
 & 75.08 \basex{0.0} & 76.06 \basex{0.0}
 & 78.99 \basex{0.0} & 79.13 \basex{0.0}
\\

& OPERA \citep{opera2024cvpr}
 & \underline{88.27} \up{4.5} & \underline{87.52} \up{4.6}
 & \underline{83.07} \up{4.9} & \underline{82.93} \up{4.6}
 & \underline{80.77} \up{5.7} & \underline{81.05} \up{5.0}
 & \underline{84.04} \up{5.1} & \underline{83.83} \up{4.7}
\\

& ICD \citep{wang2024mitigating}
 & 84.90 \up{1.2} & 84.22 \up{1.3}
 & 78.37 \up{0.2} & 78.81 \up{0.4}
 & 75.97 \up{0.9} & 76.93 \up{0.9}
 & 79.75 \up{0.8} & 79.99 \up{0.9}
\\

& VCD \citep{vcd2024cvpr}
 & 86.65 \up{2.9} & 86.99 \up{4.0} 
 & 80.73 \up{2.6} & 82.24 \up{3.9}
 & 76.09 \up{1.0} & 78.78 \up{2.7}
 & 81.16 \up{2.2} & 82.67 \up{3.5}
\\

& {\textbf{InEx (ours)}}
 & \sethlcolor{lightpink}\hl{\textbf{90.12}} \up{6.4} 
 & \sethlcolor{lightpink}\hl{\textbf{90.32}} \up{7.4}
 & \sethlcolor{lightpink}\hl{\textbf{86.37}} \up{8.2}
 & \sethlcolor{lightpink}\hl{\textbf{85.98}} \up{7.6}
 & \sethlcolor{lightpink}\hl{\textbf{82.91}} \up{7.8}
 & \sethlcolor{lightpink}\hl{\textbf{83.82}} \up{7.8}
 & \sethlcolor{lightpink}\hl{\textbf{86.72}} \up{7.7}
 & \sethlcolor{lightpink}\hl{\textbf{86.81}} \up{7.7}
\\  
\bottomrule
\end{tabular}
}
\caption{Performance on POPE benchmark using LLaVA-1.5-7B. The best results are in \sethlcolor{lightpink}\hl{\textbf{blue}}. We report accuracy and F1-score under three settings, e.g., Random, Popular, Adversarial, and Average, to show the robustness of the different methods directly. } 
\label{tab:res_pope}
\end{table*}

\subsection{Ex: Cross-modal Multi-Agent Collaboration}
Inspired by real-world decision-making, we introduce \textbf{Ex}, an external cross-modal multi-agent collaboration mechanism for alternating verification and refinement (see Algorithm~\ref{alg:inex}).
The process begins with the decision agent $\mathcal{A}_d$ generating an initial response $\mathbf{y}_0$ based on the input image $v$ and text query $\mathbf{x}$ (Line~2).
The textual self-reflection agent $\mathcal{A}_t$ verifies whether the response $\mathbf{y}_0$ is supported by the dense caption $\mathbf{c}$, serving as a validation filter.
If it is supported, the process returns $\mathbf{y}\leftarrow\mathbf{y}_0$. 
Otherwise, $\mathcal{A}_t$ sends textual feedback $\mathbf{f}_t$ to $\mathcal{A}_d$, prompting it to refine the response through introspection and initiate cross-modal verification.
To mimic a rigorous human self-reflection process, 
$\mathcal{A}_t$ performs verification and generates feedback through multi-perspective reflection and ensemble aggregation.
\begin{algorithm}[ht]
\caption{\textbf{The InEx in VQA}}\label{alg:inex}

\KwIn {\small{image $v$, text query $\mathbf{x}$, decision agent $\mathcal{A}_d$, visual self-reflection agent $\mathcal{A}_v$, textual self-reflection agent $\mathcal{A}_t$, image editing agent $\mathcal{A}_i$}}
\KwOut{Answer $\mathbf{y}$}
\textbf{Initialisation:} $\mathbf{c}$ \tcp*{\scriptsize{dense caption for $v$}}

$\mathbf{y}_0 = \mathcal{A}_d(v, \mathbf{x})$ \tcp{\scriptsize{initial generation}}
\If {$\mathcal{A}_t (\mathbf{c}, \mathbf{y}_0) ==$ \text{True}} {
$\mathbf{y} \leftarrow \mathbf{y}_0$ \\
\textbf{return} $\mathbf{y}$ }
$\mathbf{f}_t \leftarrow \mathcal{A}_t (\mathbf{c} , \mathbf{y}_i)$ \tcp{\scriptsize{feedback from text modality}}
\For{$i = 1$ \KwTo $I$}{

$ \mathbf{y}_i \leftarrow \mathcal{A}_d(v, \mathbf{x}, \mathbf{f}_t)$\tcp{\scriptsize{decision from text modality}}

 $v' \leftarrow \mathcal{A}_i (\mathbf{y}_i, v)$ \tcp{\scriptsize{image editing}}

 \If{ $\mathcal{A}_v (v,v') == \text{True}$}{
 $\mathbf{y} \leftarrow \mathbf{y}_i$
 
 \textbf{return} $\mathbf{y}$ \tcp{\scriptsize{verification from vision modality}}}

$\mathbf{f}_v\leftarrow\mathcal{A}_v(v, v^{\prime})$\tcp{\scriptsize{feedback from vision modality}}

$\mathbf{y}_i \leftarrow \mathcal{A}_d\left(v, \mathbf{x}, \mathbf{f}_v\right)$\tcp{\scriptsize{decision from vision modality}}

\If{$\mathcal{A}_t (\mathbf{c},\mathbf{y}_i) == \text{True}$}{
$\mathbf{y} \leftarrow \mathbf{y}_i$\\
\textbf{return} $\mathbf{y}$\tcp{\scriptsize{verification from text modality}}
}

$\mathbf{f}_t \leftarrow \mathcal{A}_t (\mathbf{c} , \mathbf{y}_i)$ \tcp{\scriptsize{feedback from text modality}}
$t \leftarrow t+1$
}
\end{algorithm}
 
\noindent 
The multi-perspective approach enables 
$\mathcal{A}_t$ to focus on different aspects mentioned in the text, such as actions, objects, and colors.
The ensemble mechanism repeats the verification and feedback generation multiple times. 
The verification results and their supporting reasoning evidence are then aggregated 
through a self-aggregation process to form $\tilde{\mathbf{y}}$. 
If $\tilde{\mathbf{y}} \neq \mathbf{y}'$, the aggregated evidence is returned as $\mathbf{f}_t$. 
With the additional information $\mathbf{f}_t$, $\mathcal{A}_d$ generates a revised response $\mathbf{y}_i$ (Line~8), which is then verified through the visual modality.
In Line~9, $\mathbf{y}_i$ and the original image $v$ are input to the image editing agent $\mathcal{A}_i$, which performs image augmentation to produce a new image $v'$.
In Line 10, the visual self-reflection agent $\mathcal{A}_v$ compares $v'$ with the original image $v$. 
If the similarity exceeds the threshold $\gamma_\text{CLIP}$, 
$\mathbf{y}\leftarrow\mathbf{y}_i$ is returned.
Otherwise, in Line~13, $\mathcal{A}_v$ generates visual feedback $\mathbf{f}_v$ and sends it to $\mathcal{A}_d$. 
The decision agent then revises its response based on visual feedback $\textbf{f}_v$, which is subsequently verified by $\mathcal{A}_t$.
This process repeats until a final response $\mathbf{y}$ is returned or a preset iteration limit $I$ is reached. Lines 12 and 17 together define the cross-modal consensus.

The design of Ex is guided by four core principles: (i) We assume that a response can be verified through a different modality (Lines 10 and 15), serving as a safeguard against hallucination. This mirrors how humans validate information by consulting multiple independent sources or by seeking external verification from individuals with different expertise and backgrounds. (ii) During text-based verification (Line 15), the decision must be consistent with the dense caption set $\mathbf{c}$, serving as a grounded textual representation of the visual content. This consistency ensures that the response is grounded in observable evidence, rather than being fabricated. (iii) For decisions inferred from text, we assess their validity by editing the image based on the generated output (Line 10). If the edited image remains coherent with the original, it confirms that the decision is anchored in visual reality rather than speculative reasoning. (iv) Cross-modal feedback is incorporated into $\mathcal{A}_d$ to enhance its internal introspective reasoning, enabling it to generate a more reliable response, analogous to how humans iteratively refine their decisions by drawing on diverse sources of information.





In summary, \textbf{InEx} unifies internal introspective reasoning with external cross-modal collaboration to jointly verify and improve responses in an iterative loop, jointly enhancing response reliability and mitigating hallucination.

\subsection{Theoretical Analysis} \label{sec:Theoretical}

To further illustrate the principles behind \textbf{InEx} and to comprehensively justify its effectiveness in mitigating hallucinations and ensuring robust performance, we present three theorems below. Let $\mathbf{H}$ denote the hidden states derived from the final original Transformer layer, and let $\hat{\mathbf{H}}$ represent the updated states obtained after injecting visual evidence $\mathbf{z}$ through Self-Introspective Visual Augmentation and Enhanced Logits.

\begin{theorem}
\textbf{(Mutual Information Increase)} \textit{Mutual Information (MI) quantifies the amount of information one variable contains about another. InEx increases the MI between $\hat{\mathbf{H}}$ and $\mathbf{z}$:}
\begin{equation}
    I(\hat{\mathbf{H}}; \mathbf{z}) \geq I(\mathbf{H}; \mathbf{z}).
\end{equation}
\label{thm1}
\end{theorem}

\begin{theorem}
\textbf{(Conditional Entropy Reduction)} \textit{Let $\boldsymbol{y}$ be the target output dependent on hidden states. An increase in MI leads to a corresponding reduction in the conditional entropy of the predicted output:} 
\begin{equation}
    E(\boldsymbol{y} \mid \hat{\mathbf{H}}) \leq E(\boldsymbol{y} \mid \mathbf{H}).
\end{equation}
\label{thm2}
\end{theorem}

\begin{theorem}
\textbf{(IB Objective Improvement)} \textit{Under the Information Bottleneck (IB) framework, InEx with Self-Introspective Visual Augmentation and Enhanced Logits optimizes the objective function by reducing the IB loss:} 
\begin{equation}
    \mathcal{L}(\hat{\mathbf{H}}) \leq \mathcal{L}(\mathbf{H}),
\end{equation}
\textit{where} $\mathcal{L}(\mathbf{H}) = I(\mathbf{H}; \boldsymbol{x}) - \beta I(\mathbf{H}; \boldsymbol{y})$, \textit{$\boldsymbol{x}$ denotes the text query embedding and $\beta$ is a trade-off parameter.}
\label{thm3}
\end{theorem}

Our theoretical foundation builds on the Data Processing Inequality (DPI) \citep{cover1991entropy} and the contraction properties of stochastic mappings in deep networks, as discussed in the IB literature \citep{achille2018information}. By increasing mutual information with visual input and reducing uncertainty in hidden states, InEx effectively mitigates hallucination.
\begin{figure*}[!t]
\centering
\includegraphics[width=1.05\linewidth]{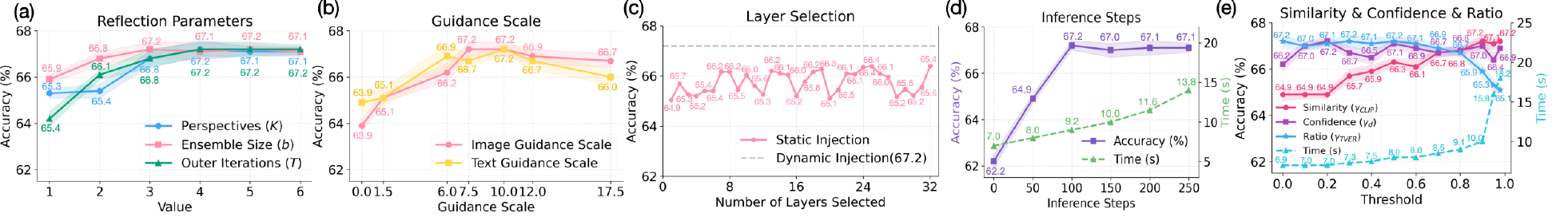}
\vspace{-3mm}
\caption{InEx Parameter Analysis on MMBench using LLaVA-1.5-7B.}
\label{fig:parameter}
\vspace{-2mm}
\end{figure*}

\vspace{-1mm}
\section{Experiments}
\label{sec:experiments}

\vspace{-1mm}
\subsection{Experiment Setup}
\textbf{Datasets.} 
To rigorously assess the effectiveness of InEx, we conduct a comprehensive set of experiments across a diverse set of benchmarks, as detailed below:
\begin{itemize}
\setlength\itemsep{0pt}
\setlength\parsep{0pt}
\setlength\topsep{0pt}
    \item \textbf{Hallucination Benchmarks:} POPE benchmark~\cite{pope}, CHAIR~\cite{rohrbach2018object}, HallusionBench~\cite{hallubench}.
    \item \textbf{General-purpose Benchmarks:} VizWiz-VQA \cite{gurari2018vizwiz}, MME~\cite{fu2023mme}, MMBench~\cite{liu2024mmbench}, MM-Vet~\cite{mmvet}, LLaVA-Bench (in-the-wild)~\cite{llava}. 
\end{itemize}

\noindent \textbf{Baselines and Models:} InEx is evaluated against three representative hallucination mitigation baseline methods: OPERA~\citep{opera2024cvpr}, ICD~\citep{wang2024mitigating}, and VCD~\citep{vcd2024cvpr}. The evaluation is conducted using Qwen-VL~\citep{Qwen-VL}, LLaVA-1.5~\citep{llava}, and GLM-4V~\citep{glm2024chatglm},  covering 7B to 10B scales.

\noindent \textbf{Implementation Details:}
We set $\gamma_\text{TVER} = 0.55$, $\gamma_\text{d} = 0.2$, $\gamma_\text{CLIP} = 0.9$, $\alpha_1 = 1$, and $\alpha_2 = 1$. Self-reflection uses temperature-controlled decoding with a value of 0.7, combined with ensemble aggregation across three outputs and up to four specialized perspectives. The editing agent is implemented via IC-Edit~\cite{zhang2025ICEdit}. All experiments are conducted on NVIDIA H100 GPUs with FP16 precision. 

\begin{table}[t]
\centering
\renewcommand{\arraystretch}{1.2}
\resizebox{1.0\linewidth}{!}{
\belowrulesep=0pt
\aboverulesep=0pt
\begin{tabular}{l|lccccc}
\toprule
\multirow{2}{*}[-0.5ex]{{\textbf{Model}}} 
 & \multirow{2}{*}[-0.4ex]{{\textbf{Methods}}} 
 & \multicolumn{1}{c}{{~\textbf{HallusionBench}}} 
 & \multicolumn{1}{c}{{~\textbf{LLaVA-Bench}}}  
 & \multicolumn{3}{c}{{\textbf{CHAlR}}} 
 \\ 
\cmidrule(l){3-3} \cmidrule(l){4-4} \cmidrule(l){5-7}  
&  
& {Accuracy} $\uparrow$
& {Accuracy} $\uparrow$
& CHAIR$_S$ $\downarrow$ 
& CHAIR$_I$ $\downarrow$ 
& {Recall} $\uparrow$
\\ 
\midrule

\multirow{2}{*}[-4ex]{{LLaVA-1.5-7B}}
& {\textemdash} 
   & \underline{41.5} \basex{0.0} 
   & \underline{63.4} \basex{0.0} 
   & 50.0 \basex{0.0} 
   & {15.4} \basex{0.0} 
   & 77.1 \basex{0.0} 
   \\

& OPERA  
   & {41.2} \downbad{0.3} 
   & {59.8} \downbad{3.6} 
   & \underline{47.9} \downbad{2.1} 
   & \underline{14.5} \downbad{0.9} 
   & 76.8 \downbad{0.3}
   \\

& ICD  
   & 38.2 \downbad{3.3}
   & 49.8 \downbad{13.}
   & 56.2  \up{6.2}
   & {16.3} \up{0.9}
   & 76.3 \downbad{0.8}
   \\

& VCD 
   & 41.1  \downbad{0.4} 
   & 59.1 \downbad{4.3}
   & 48.6 \downbad{1.4}
   & 14.9 \down{0.5}
   & \underline{77.3} \up{0.2}
   \\

& {\textbf{InEx (Ours)}}
   & \sethlcolor{lightpink}\hl{\textbf{44.2}} \up{2.7}
   & \sethlcolor{lightpink}\hl{\textbf{66.5}} \up{3.1}
   & \sethlcolor{lightpink}\hl{\textbf{45.1}} \downbad{4.9}
   & \sethlcolor{lightpink}\hl{\textbf{13.3}} \downbad{4.1}
   & \sethlcolor{lightpink}\hl{\textbf{83.2}} \up{6.1}
   \\ \bottomrule
\end{tabular}
}
\caption{Performance evaluation on HallusionBench, LLaVA-Bench, and CHAlR. Accuracy means global accuracy.}
\label{tab:all2}
\end{table}

\subsection{Results on Hallucination Benchmarks}

Table~\ref{tab:res_pope} shows hallucination results on POPE, where InEx consistently outperforms all baselines across three sub-datasets and question settings (random, popular, adversarial). On the MSCOCO subset, it boosts LLaVA-1.5-7B accuracy from 79.83 to 88.73(+ 8.9\%). 
InEx also achieves the best results on HallusionBench and CHAIR across multiple metrics, as shown in Table~\ref{tab:all2}. 

\vspace{-1mm}
\subsection{Results on General-purpose Benchmarks}
We further evaluate InEx on general-purpose benchmarks. As shown in Table~\ref{tab:all2}, InEx is the only method applied to LLaVA-1.5 on LLaVA-Bench, improving accuracy from 63.4\% to 66.5\%. In Table~\ref{tab:all3}, InEx consistently outperforms existing methods for LLaVA-1.5, Qwen-VL, and GLM-4V across all datasets and metrics.
For LLaVA-1.5, it achieves the highest score on MME-Hall (673.3), with strong results on object-level (Existence: 199.3, Count: 157.7) and attribute-level (Position: 137.0, Color: 179.3) metrics. It also leads on MM-Vet (36.00), VizWiz (53.80), and MMBench (67.17). InEx also demonstrates notable improvements for Qwen-VL and GLM-4V, particularly on MME-Hall, achieving scores of 677.3 (+59) and 732.3 (+35), respectively. These results highlight InEx’s robust capability in object and attribute understanding. Notably, OPERA does not support Qwen-VL and GLM-4V. 

\vspace{-1mm}
\subsection{Ablation Studies}

We conduct ablation studies to analyze the individual contributions of \textbf{In} and \textbf{Ex} on seven datasets. For Ex, we evaluate both the textual and visual self-reflection components. Table~\ref{tab:ablation} shows that each component contributes to performance gains when used individually.
However, the best performance is achieved when all three are combined, confirming the complementary effects of introspective reasoning and cross-modal verification in mitigating hallucination.

\begin{table}[!ht]
\centering
\resizebox{\linewidth}{!}{
\begin{tabular}{cccccccccc}
\toprule
\textbf{In} & \textbf{Textual} & \textbf{Visual}  & \textbf{POPE} & \textbf{CHAIR} & \textbf{VizWiz} & \textbf{MME} & \textbf{MMBench} & \textbf{MM-Vet} & \textbf{LLaVA-Bench} 
\\
\midrule
\usym{1F5F6} & \usym{1F5F6} & \usym{1F5F6}  & 79.83 & 77.10  & 50.00 & 643.3  & 62.80 & 31.10 & 64.80  \\
\usym{2714}  & \usym{1F5F6} & \usym{1F5F6}  & 86.43 & 79.40 & 51.30  & 648.5 & 63.99 & 32.60 & 65.40   \\
\usym{1F5F6} & \usym{2714}  & \usym{1F5F6}   & 83.20 & 78.20 & 45.90  & 641.5 & 63.08 & 32.10 & 64.90   \\
\usym{1F5F6} & \usym{1F5F6} & \usym{2714}  & 85.20 & 80.20 & 49.70 & 645.5 & 63.71 & 31.70 & 65.30   \\
\usym{2714} & \usym{2714}  & \usym{1F5F6}  & 86.39 & 81.90 & 51.95 & 649.3 & 64.92 & 33.20 & 65.20   \\
\usym{2714} & \usym{1F5F6} & \usym{2714}  & \underline{87.77} & \underline{82.50} & \underline{52.25} & 645.8 & \underline{65.22} & \underline{35.10} & \underline{66.10}  \\
\usym{1F5F6}  & \usym{2714}  & \usym{2714} & 86.93 & 82.40 & 51.90  & \underline{650.5} & 64.97 & 33.00 & 65.70   \\
\rowcolor{lightpink}
\usym{2714}  & \usym{2714}  & \usym{2714}  & \textbf{88.73} & \textbf{83.20} & \textbf{53.80} & \textbf{653.8}  & \textbf{67.17} & \textbf{36.00} & \textbf{66.50} \\
\bottomrule
\end{tabular}
}
\caption{Ablation studies on LLaVA-1.5-7B.}
\label{tab:ablation}
\end{table}

\begin{table*}[ht]
\centering
\belowrulesep=0pt
\aboverulesep=0pt
\vspace{-3mm}
\renewcommand{\arraystretch}{1.2}
\resizebox{1.0\linewidth}{!}{
\begin{tabular}{l|lcccccccc}
\toprule 
\multirow{2}{*}{\textbf{Models}} & \multirow{2}{*}{\textbf{Methods}} & \textbf{MME-Hall}                          & \multicolumn{2}{c}{\textbf{Object-Level}}                                                 & \multicolumn{2}{c}{\textbf{Attribute-Level}}                                           & \textbf{MM-Vet}                           & \textbf{Vizwiz}                           & \textbf{MMBench}                          \\ \cmidrule(l){3-3} \cmidrule(l){4-5} \cmidrule(l){6-7} \cmidrule(l){8-8} \cmidrule(l){9-9} \cmidrule(l){10-10}
                                     &                                   & {Total} $\uparrow$                         & {Existence} $\uparrow$                       & {Count} $\uparrow$                         & {Position} $\uparrow$                     & {Color} $\uparrow$                         & Accuracy $\uparrow$                       & Accuracy $\uparrow$                       & Accuracy $\uparrow$                       \\ \hline
\multirow{5}{*}{\textbf{LLaVA-1.5-7B}}  &    \textemdash                      & {643.3} \basex{0.0}                        & {190.0} \basex{0.0}                          & \underline{155.0} \basex{0.0}                          & {128.3} \basex{0.0}                       & \underline{170.0} \basex{0.0}                          & 31.10 \basex{0.0}                         & 50.00 \basex{0.0}                         & \underline{62.80} \basex{0.0}                       \\
                                     & OPERA \citep{opera2024cvpr}\;     & {610.0} \downbad{33.}                      & \underline{195.0} \up{5.0}                             & 128.3 \downbad{26.}                        & {121.7} \downbad{6.6}                     & 165.0 \downbad{5.0}                        & \underline{32.00} \up{0.9}                          & \underline{50.76} \up{0.8}                          & \underline{62.80} \basex{0.0}                       \\
                                     & ICD \citep{wang2024mitigating}    & 583.3 \downbad{60.}                        & 185.0 \downbad{5.0}                          & 130.0 \downbad{25.}                        & {121.7} \downbad{6.6}                     & 146.7 \downbad{23.}                        & 25.90 \downbad{5.2}                       & 37.62 \downbad{12.}                       & 39.78 \downbad{23.}                       \\
                                     & VCD \citep{vcd2024cvpr}           & \underline{648.3} \up{5.0}                           & {190.0} \basex{0.0}                          & \underline{155.0} \basex{0.0}                          & \underline{133.3} \up{5.0}                            & \underline{170.0} \basex{0.0}                          & 30.20 \downbad{0.9}                       & 44.90 \downbad{5.1}                       & 54.21 \downbad{8.6}                       \\
                                     & \textbf{InEx (Ours)}                         & \sethlcolor{lightpink}\hl{\textbf{673.3}} \up{30.} & \sethlcolor{lightpink}\hl{\textbf{199.3}} \up{9.3} & \sethlcolor{lightpink}\hl{\textbf{157.7}} \up{2.7}  & \sethlcolor{lightpink}\hl{\textbf{137.0}} \up{8.7} & \sethlcolor{lightpink}\hl{\textbf{179.3}} \up{9.3}  & \sethlcolor{lightpink}\hl{\textbf{36.00}} \up{4.9} & \sethlcolor{lightpink}\hl{\textbf{53.80}} \up{3.8} & \sethlcolor{lightpink}\hl{\textbf{67.17}} \up{4.4} \\ \cmidrule(l){1-2}
\multirow{4}{*}{\textbf{Qwen-VL-10B}}    & \textemdash                           & 618.3 \basex{0.0}                          & \underline{175.0} \basex{0.0}                          & \underline{140.0} \basex{0.0}                        & 123.3 \basex{0.0}                         & 180.0 \basex{0.0}                          & \underline{49.00} \basex{0.0}                       & \underline{66.05} \basex{0.0}                       & \underline{56.53} \basex{0.0}                       \\
                                     & ICD \citep{wang2024mitigating}    & 616.7 \downbad{1.7}                        & 170.0 \downbad{5.0}                          & {138.3} \downbad{1.7}                      & \underline{148.3} \up{25.}                          & 160.0 \downbad{20.}                        & 31.70 \downbad{17.}                       & 29.37 \downbad{36.}                       & 13.32 \downbad{43.}                       \\
                                     & VCD \citep{vcd2024cvpr}           & \underline{648.3} \up{30.}                           & \underline{175.0} \basex{0.0}                          & 130.0 \downbad{10.}                        & \sethlcolor{lightpink}\hl{\textbf{153.3}} \up{30.} & \underline{190.0} \up{10.}                             & {34.60} \downbad{14.}                     & {34.54} \downbad{31.}                     & {39.18} \downbad{17.}                     \\
                                     & \textbf{InEx (Ours)}                           & \sethlcolor{lightpink}\hl{\textbf{677.3}} \up{59.} & \sethlcolor{lightpink}\hl{\textbf{189.0}} \up{13.}    & \sethlcolor{lightpink}\hl{\textbf{151.0}} \up{11.0} & 143.3 \up{20.}                           & \sethlcolor{lightpink}\hl{\textbf{194.0}} \up{14.} & \sethlcolor{lightpink}\hl{\textbf{52.30}} \up{3.3} & \sethlcolor{lightpink}\hl{\textbf{68.56}} \up{2.5} & \sethlcolor{lightpink}\hl{\textbf{59.74}} \up{3.2} \\ \cmidrule(l){1-2}
\multirow{4}{*}{\textbf{GLM-4V-9B}}    & \textemdash                           & 697.3 \basex{0.00}                          & \underline{195.0} \basex{0.0}                          & {165.0} \basex{0.0}                        & 153.3 \basex{0.0}                         & \underline{184.0} \basex{0.0}                          & \underline{63.40} \basex{0.0}                       & \underline{57.39} \basex{0.0}                       & \underline{82.39} \basex{0.0}                       \\
& ICD \citep{wang2024mitigating}    &   680.0 \downbad{17.3}                          & {185.0} \downbad{10.}                          & {160.0} \downbad{5.0}                        & \underline{170.0} \up{16.7}                         & 165.0 \downbad{19.}                          &  59.40 \downbad{4.0}                       & 48.07 \downbad{9.4}                       & 78.01 \downbad{4.4}                       \\
& VCD \citep{vcd2024cvpr}   &  \underline{714.0} \up{16.7}   & \underline{195.0} \basex{0.0}    & \underline{173.3} \up{8.3}       & 160.7 \up{7.4}       & \sethlcolor{lightpink}\hl{\textbf{185.0}} \up{1.0}         & {57.70} \downbad{5.7} & {50.01} \downbad{7.4}  
& {80.58} \downbad{1.8}   \\

& \textbf{InEx (Ours)}                         
& \sethlcolor{lightpink}\hl{\textbf{732.3}} \up{35.0}   & \sethlcolor{lightpink}\hl{\textbf{199.0}} \up{4.0}   
& \sethlcolor{lightpink}\hl{\textbf{175.0}} \up{15.}       & \sethlcolor{lightpink}\hl{\textbf{173.3}} \up{20.}       & \sethlcolor{lightpink}\hl{\textbf{185.0}} \up{1.0}         & 
\sethlcolor{lightpink}\hl{\textbf{68.80}} \up{5.4} & \sethlcolor{lightpink}\hl{\textbf{60.36}} \up{3.0} & \sethlcolor{lightpink}\hl{\textbf{85.65}} \up{3.3} \\ \bottomrule
\end{tabular}
}
\vspace{-1mm}
\caption{Performance evaluation on MME Hallucination subset, MM-Vet, Vizwiz, and MMBench. The best results are in \sethlcolor{lightpink}\hl{blue}. }
\label{tab:all3}
\vspace{-2mm}
\end{table*}

\vspace{-1mm}
\subsection{Different Image Editing Model}
\begin{figure}[t]
\centering
\includegraphics[width=\linewidth]{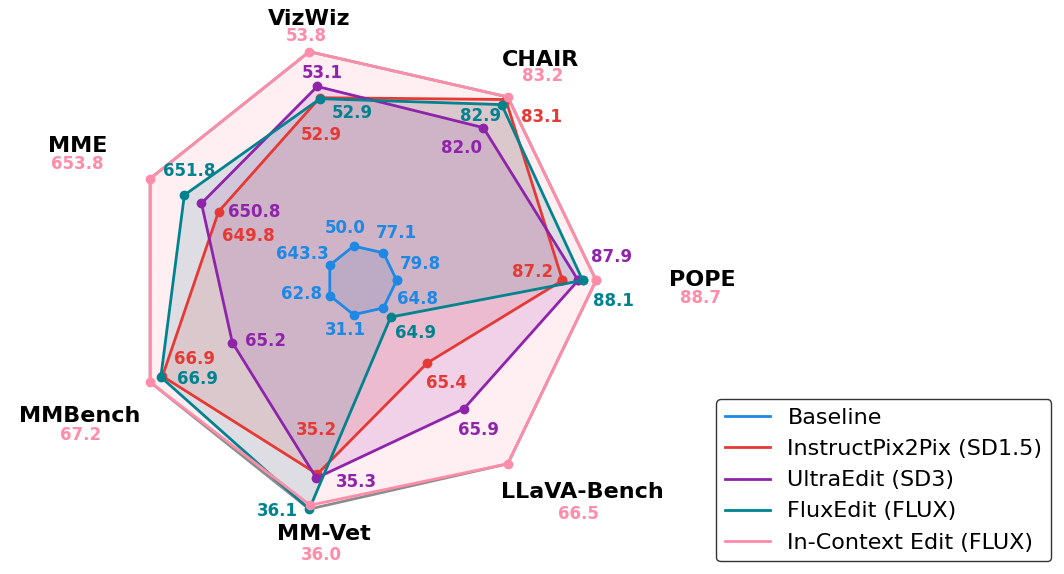}
\vspace{-2mm}
\caption{We conduct an ablation study to analyze the effect of 4 different image editing models using LLaVa-1.5-7B.}
\label{fig:image_ab}
\vspace{-4mm}
\end{figure}

Furthermore, the image editing model is a key component in the design of external verification. To evaluate its impact, we experiment with four image editing models with different architectures (diffusion-based and FLUX-based). As shown in Figure~\ref{fig:image_ab}, InEx with In-Context Edit achieves the best overall performance. Nevertheless, InEx consistently improves upon the original model across all editing variants.

\subsection{Parameter Analysis}

\begin{table}[!ht]
\centering
\resizebox{\linewidth}{!}{
\begin{tabular}{lcccc}
\toprule
\textbf{Method} & \textbf{Mean (\%)} & \textbf{StdDev} & \textbf{$t$-test $p$} & \textbf{Wilcoxon $p$}  \\
\midrule
OPERA  \citep{opera2024cvpr} & 84.14  & 0.62  & $8.72 \times 10^{-25}$ & $9.54 \times 10^{-7}$ \\
ICD \citep{wang2024mitigating} &  82.97 &  0.41 & $2.85 \times 10^{-37}$ & $9.54 \times 10^{-7}$ \\
VCD \citep{vcd2024cvpr} & 82.60  & 0.54  & $7.95 \times 10^{-30}$ & $9.54 \times 10^{-7}$ \\
\rowcolor{lightpink}\textbf{InEx} & 88.73 & 0.33 & - & - \\
\bottomrule
\end{tabular}
}
\vspace{-2mm}
\caption{
Statistical comparison on \textsc{POPE} benchmark. \textbf{StdDev} denotes standard deviation. Paired one-sided $t$-test and Wilcoxon signed-rank test $p$-values are reported ($\alpha = 5\%$).}
\label{tab:significance}
\vspace{-5mm}
\end{table}

\noindent \textbf{Agents:} 
For the text self-reflection agent, we evaluate two key parameters: perspective, which denotes the number of different viewpoints used to verify a decision and generate feedback, and ensemble size, which indicates how many times the process is repeated. Both parameters lead to stable performance gains up to a value of four, effectively capturing diverse textual evidence through repetition. For the external collaboration process, we analyze the effect of the maximum number of iterations and observe that performance stabilizes when this value reaches four. This suggests that additional iterations yield diminishing returns, indicating that the verification mechanism has reached its effective capacity. All corresponding results are shown in Figure~\ref{fig:parameter}(a). Figure~\ref{fig:parameter}(b) shows the impact of image and text guidance scales, two key parameters of the image editing agent. Image guidance peaks at a scale of 7.5, suggesting that moderate constraints help preserve visual cues, while higher values introduce artifacts that reduce alignment. In contrast, text guidance improves performance steadily up to 10, where stronger semantic signals better align generated images with intended content. Furthermore, Figure~\ref{fig:parameter}(d) illustrates that increasing the number of inference steps for the editing agent improves accuracy up to 100 steps by reducing noise and enhancing clarity. Beyond this point, however, performance gains taper off while computational cost continues to increase.

\noindent \textbf{Threshold:}
InEx includes three thresholds: $\gamma_\text{TVER}$, $\gamma_\text{d}$, and $\gamma_\text{CLIP}$. 
As shown in Figure~\ref{fig:parameter}(e), increasing $\gamma_\text{TVER}$ results in decreased accuracy, highlighting the importance of TVER in triggering introspective internal reasoning. A lower $\gamma_\text{TVER}$ makes the system more sensitive to ambiguity and more likely to activate introspection. In contrast, increasing $\gamma_\text{CLIP}$ enhances accuracy by strengthening the verification capacity of the visual self-reflection agent, further highlighting the benefits of cross-modal alignment. The final threshold, $\gamma_\text{d}$, determines the decoding mode. Accuracy fluctuates as $\gamma_\text{d}$ changes, indicating that collaborative and contrastive decoding modes can complement each other effectively.

\noindent \textbf{Layer selection:}
Self-introspective visual augmentation requires selecting a Transformer layer to inject visual features. Figure~\ref{fig:parameter}(c) compares fixed-layer selection with dynamic selection informed by the TVER value. The dynamic approach outperforms fixed-layer selection, demonstrating the benefit of adapting injection depth based on reasoning uncertainty.

Notably, InEx consistently enhances performance across all parameter configurations, as shown in Figure~\ref{fig:parameter}.
 
\subsection{Statistical Significance Study}
We conduct 20 independent runs with different random seeds and compare our method against baseline models under identical settings. To assess statistical significance, we perform paired one-sided $t$-tests and Wilcoxon signed-rank tests. 
The null hypothesis ($\mathbf{H}_0$) states that InEx performs equally or worse than the baseline, while the alternative hypothesis ($\mathbf{H}_1$) states that our method performs better.
As shown in Table~\ref{tab:significance}, our method achieves the highest mean score (88.73 $\pm$0.33), compared to OPERA (84.14 $\pm$0.62), ICD (82.97 $\pm$0.41), and VCD (82.60 $\pm$0.54). 
All obtained $p$-values are significantly below the threshold $\alpha = 0.05$ (e.g., $t$-test: $8.72 \times 10^{-25}$ against OPERA), thereby allowing us to confidently reject $\mathbf{H}_0$ in favor of $\mathbf{H}_1$.
These results confirm that our method consistently and significantly outperforms all baselines.

\begin{figure}[!t]
\centering
\includegraphics[width=\linewidth]{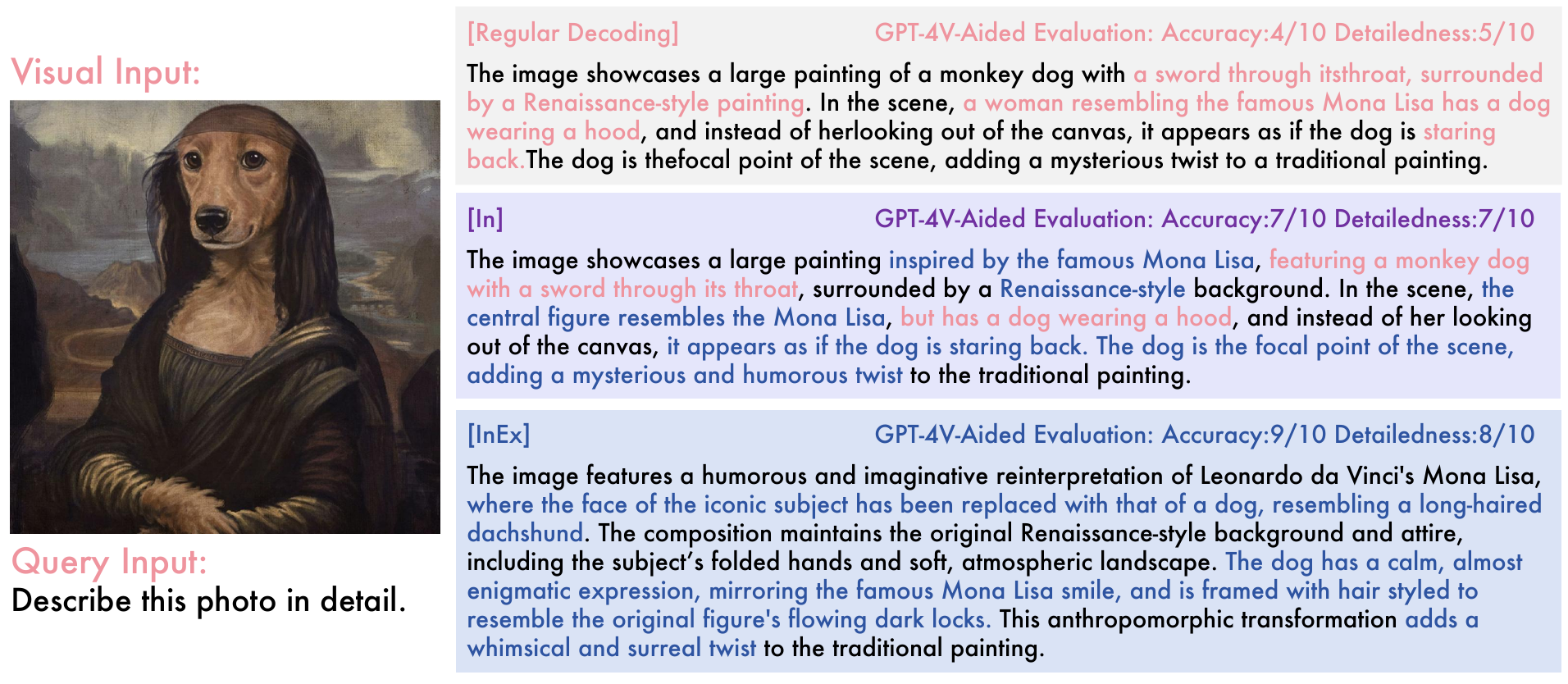}
\vspace{-3mm}
\caption{Case study on LLaVA-Bench comparing responses from standard decoding and our method using LLaVA-1.5. GPT-4V-aided evaluations are shown, with hallucinated and accurate content highlighted in red and blue.}
\label{fig:case_study1}
\vspace{-1mm}
\end{figure}

\subsection{Cases Study}
As shown in Figure \ref{fig:case_study1}, InEx generates more accurate and grounded image descriptions. When presented with a stylized parody of the Mona Lisa, regular decoding failed to identify key visual transformations, resulting in vague and partially incorrect content (accuracy: 4/10, detailedness: 5/10). For instance, it misinterpreted the central subject and overlooked stylistic nuances. In contrast, InEx accurately recognized the dog-faced anthropomorphic transformation of the Mona Lisa, preserved essential compositional elements such as the folded hands and Renaissance-style background, and captured the whimsical tone of the image (accuracy: 9/10, detailedness: 8/10), highlighting InEx’s effectiveness in mitigating hallucination and enhancing visual-text alignment. 

\vspace{-1mm}
\section{Limitation and Future Work}
InEx is designed for MLLMs operating in the text and vision modalities. However, it currently lacks effective support for the audio modality. To extend InEx to more modalities, we plan to add an audio self-reflection agent in the future, enabling cross-modal verification and enhancing MLLMs that process audio alongside vision and text.

\section{Conclusion}
We introduced \textbf{InEx}, a training-free multi-agent framework that reduces hallucination by combining internal introspection (\textbf{In}) with external cross-modal collaboration (\textbf{Ex}). Experiments show that InEx consistently surpasses strong baselines across diverse hallucination and general-purpose benchmarks. Ablation studies confirm the contributions of both In and Ex. Beyond mitigating hallucination, InEx highlights how moving from single-agent reasoning to coordinated multi-agent interaction enables more reliable, autonomous behavior with minimal human intervention.

\bibliography{aaai2026}
\end{document}